\documentclass{ecai} 

\usepackage{latexsym}
\usepackage{amssymb}
\usepackage{amsmath}
\usepackage{amsthm}
\usepackage{booktabs}
\usepackage{enumitem}
\usepackage{graphicx}
\usepackage{color}

\newcommand{\BibTeX}{B\kern-.05em{\sc i\kern-.025em b}\kern-.08em\TeX}

\usepackage{todonotes}

\usepackage{tikz}
\usetikzlibrary{arrows,automata,shapes.geometric}

\usepackage{algorithm}
\usepackage{algorithmicx,algpseudocode}

\begin{document}


\begin{frontmatter}


\paperid{123} 


\title{Cloud Kitchen: Using Planning-based Composite AI\\to Optimize Food Delivery Processes}


\author{\fnms{Slavom\'{i}r}~\snm{\v{S}vanc\'{a}r}}
\author{\fnms{Luk\'{a}\v{s}}~\snm{Chrpa}}
\author{\fnms{Filip}~\snm{Dvo\v{r}\'{a}k}}
\author{\fnms{Tom\'{a}\v{s}}~\snm{Balyo}}

\address{Filuta AI, Inc., 1606 Headway Cir STE 9145, Austin, TX 78754, United States}


\begin{abstract}
The global food delivery market provides many opportunities for AI-based services that can improve the efficiency of feeding the world. This paper presents the Cloud Kitchen platform as a decision-making tool for restaurants with food delivery and a simulator to evaluate the impact of the decisions. The platform contains a Technology-Specific Bridge (TSB) that provides an interface for communicating with restaurants or the simulator. TSB uses a planning domain model to represent decisions embedded in the Unified Planning Framework (UPF). Decision-making, which concerns allocating customers' orders to vehicles and deciding in which order the customers will be served (for each vehicle), is done via a Vehicle Routing Problem with Time Windows (VRPTW), an efficient tool for this problem. We show that decisions made by our platform can improve customer satisfaction by reducing the number of delayed deliveries using a real-world historical dataset.
\end{abstract}

\end{frontmatter}

\section{Introduction}

The global food delivery market accounted for \$190 billion in 2022 and is forecasted to grow over \$500 billion in 2032\footnote{https://www.precedenceresearch.com/online-food-delivery-market}. Such a market increase, accelerated by the recent COVID-19 pandemic, establishes (online) food delivery as a new norm in society~\cite{chai2019online,shankar2022online}. This opens a number of opportunities for AI-based services that can optimise the food delivery process by reducing costs and improving customer satisfaction. Apps such as UberEats\texttrademark~ involve courier services that can collect and deliver orders from multiple restaurants~\cite{STEEVER2019173} and hence the problem they deal with can be modelled as a (dynamic) Pickup and Delivery problem~\cite{DBLP:journals/transci/SavelsberghS95,DBLP:journals/ijon/CaiZLMLM23}.

In contrast to courier services (e.g. UberEats), we consider restaurants having their own food delivery. Benefits of having their own food delivery include not paying provisions for third-party order and delivery services and not relying only on apps or web services (e.g. being able to accept orders by phone). Hence, the problem we deal with in this paper concerns assigning food orders to vehicles and providing routing for the vehicles, i.e., deciding in which order the deliveries are made. Such a problem can be modelled as a domain-independent planning problem~\cite{ghallab2004automated}, in PDDL (Planning Domain Definition Language)~\cite{pddl} which is currently the most established language for specifying (domain-independent) planning tasks. Therefore, a wide range of different planning engines as well as tools for plan verification and plan execution monitoring evolved around PDDL. A recent effort to unify these planning engines and tools motivated the development of the Unified Planning Framework (UPF)~\cite{upf}. Although PDDL and UPF provide us a machinery for plan verification, explanation, and monitoring, the performance of existing domain-independent planners is rather poor for the problem. It is more effective to model our problem as a Vehicle Routing Problem with Time Windows (VRPTW)~\cite{kallehauge2005vehicle}, which is specifically designed for it. 
Embedding VRPTW into the UPF~\cite{upf} can keep both the efficiency of solving (by VRPTW) and the explainability of solutions~\cite{fox2017explainable}. It can be seen as a straightforward example of Composite AI~\cite{martie2023rapid}, which aims at an effective combination of different AI techniques for addressing a given problem, or, in the planning context, as an example of Planning Modulo Theories~\cite{GregoryLFB12}. Although we provide a rather trivial example of a Composite AI approach, it can be seen that the use of PDDL (a domain-independent language) and UPF (a domain-independent tool)  can provide a useful machinery for for plan monitoring and explanation while embedding VRPTW (a specific solver) into it provides an efficient way to solve our problem.

This paper introduces \emph{Cloud Kitchen}, a platform that provides a decision-making tool for restaurants (or kitchens) with food delivery. The platform is designed to run as a cloud service with a Technology-Specific Bridge (TSB) serving as its interface. TSB uses a PDDL model to represent decisions that are embedded into UPF. TSB passes information about new orders and available vehicles into a Decision-Making component, which formulates and solves VRPTW tasks. Solutions are then passed back to TSB, which interprets them as (PDDL) plans. Cloud Kitchen also includes a simulator that simulates decisions made on real historical datasets to demonstrate the usefulness of the platform for potential customers (i.e., restaurants). We empirically show on a historical dataset provided by our industrial partner that the Cloud Kitchen platform can improve customer satisfaction by considerably reducing the number of deliveries delayed by more than 10 minutes.

\section{Vehicle Routing Problem with Time Windows}
A \emph{Vehicle Routing Problem with Time Windows (VRPTW)} $T=(V,C,\mathcal{G})$ consists of a set of vehicles $V$, a set of customers $C$, and a graph $\mathcal{G}=(N,E)$ such that $|N|=|C|+2$. By convention, we denote the starting depot as $n_0$ and the returning depot as $n_{k+1}$ and the location of a customer $c_i$ as $n_i$ (assuming that $|C|=k$). For each edge $(i,j)\in E$, we denote by $\mathit{dist}(i,j)$, resp.~$\mathit{time}(i,j)$, the distance between locations $i$ and $j$, resp.~the driving time between $i$ and $j$ (including service time in $j$). Each vehicle $v\in V$ has its capacity $q_v$ and each customer $c_i\in C$ has its demand $d_i$ (representing the ``size'' of the delivery) and its \emph{time window} $[a_i,b_i]$ representing that the delivery has to be made at time $a_i$ at the earliest (the delivering vehicle might need to wait if it arrives earlier) and at time $b_i$ at the latest. We can also specify a \emph{scheduling horizon} $[a_0,b_{k+1}]$ representing when a vehicle can leave the depot ($a_0$) and the latest time the vehicle returns to the depot ($b_{k+1}$).

A \emph{solution} $\tau$ of a VRPTW $T$ is a set of triples $(C_v,\pi_v,t_v)$ for each $v\in V$ such that $C=\bigcup_{v\in V}C_v$, $C_v\cap C_{v'}=\emptyset$ (for all $v,v'\in V$ such that $v\neq v'$), $q_v\geq \sum_{c_i\in C_v}d_i$, $\pi_v$ is a path $\langle n_{v_0},n_{v_1},\dots,n_{v_j},n_{v_{j+1}}\rangle$ ($C_v=\{c_{v_1},\dots,c_{v_j}\}$, $n_{v_0}=n_0$, $n_{v_{j+1}}=n_{k+1}$), and $t_v$, representing time of delivery, is defined on the set of nodes from $\pi_v$ such that $t_v(n_{v_0})\geq a_0$, $\max(a_{v_i},t_v(n_{v_{i-1}})+time(v_{i-1},v_i))\leq t_v(n_{v_{i}})\leq b_{v_i}$ ($1\leq i\leq k+1$).
Solutions can be optimised, e.g., for the total traveled distance, or the total travel time, i.e., by minimising $\sum_{v\in V}\sum_{i=1}^{j+1}\mathit{dist}(v_{i-1},v_i)$, or $\sum_{v\in V}t_v(n_{v_{j+1}})$, respectively.

For more details about VRPTW, the interested reader is referred to the literature~\cite{kallehauge2005vehicle}.


\section{Cloud Kitchen Platform}

\begin{figure}[!t]
\centering
\begin{tikzpicture}[-triangle 60,auto,node distance=3cm,ultra thick,scale=0.92, every node/.style={transform shape}]
\tikzstyle{every state}=[ellipse,thick,minimum size=8mm,text=black,minimum width=2cm] 
     
    \node[state,align=center] (A) {Technology-Specific Bridge};
    \node[state] (D) [above left of=A] {Restaurants};
    \node[state] (E) [above right of=A] {Simulator}; 
    \node[state] (F) [below of=A] {Decision-Making Component};

\draw [solid] (A) edge node{decisions} (D);
\draw [solid] (D)[anchor=west] edge[bend right] node[anchor=east,align=center]{real-time\\ observations} (A.west);
\draw [solid] (A) edge node{decisions} (E);
\draw [solid] (E) edge[bend left] node[anchor=west,align=center]{historical\\ data} (A.east);

\draw [solid] (A) edge[bend right] node[anchor=east,align=center]{customers,\\ vehicles\\ data} (F);
\draw [solid] (F) edge[bend right] node[anchor=west,align=center]{plans} (A);

\end{tikzpicture} 
\caption{Architecture of the Cloud Kitchen platform}\label{fig:architecture}
\end{figure}
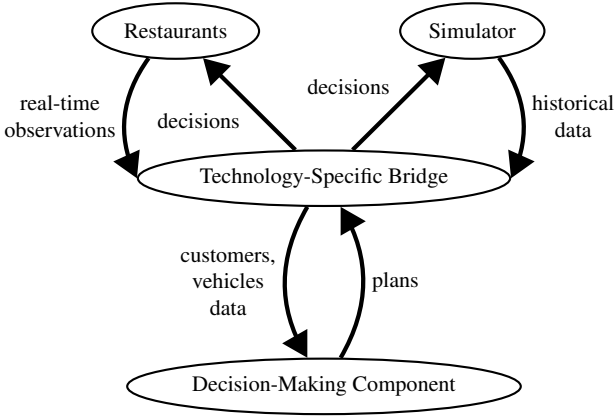


\begin{algorithm}[!t]
\begin{algorithmic}[1]
\Require Deadline extension step $\delta$ 
\While{True}
\State $C_{new}\gets$ GetNewOrders()
\State $V_{disp}\gets$ GetDispatchedVehicles()
\State $V_{new}\gets$ GetReturnedVehicles()
\State $V\gets (V\setminus V_{disp})\cup V_{new}$
\State $C\gets (C\setminus\bigcup_{v\in V_{disp}}C_v)\cup C_{new}$
\State $\mathcal{G}\gets$ GraphUpdate($\mathcal{G},C$)
\State $\mathit{delay}\gets 0$
\Repeat
\State $T\gets(V,C,\mathcal{G})$
\State $\tau\gets$ SolveVRPTW($T,\mathit{delay}$)
\If{$\tau$ is invalid}
\State $\mathit{delay}\gets\mathit{delay}+\delta$
\EndIf
\Until{$\tau$ is a valid solution of $T$}
\State SendSolution($\tau$)
\EndWhile
\end{algorithmic}
\caption{The high-level routine of the DM component}\label{alg}
\end{algorithm}

In a nutshell, \emph{Cloud Kitchen} is a platform that aims at improving the efficiency of food delivery services in restaurants. Cloud Kitchen also includes a simulator that can simulate the process of food ordering and delivery based on historical data, which can be used to demonstrate the impact of the platform on the delivery business of restaurants. 
The architecture of the Cloud Kitchen framework is depicted in Figure~\ref{fig:architecture}. The Technology-Specific Bridge provides an interface to restaurants (the framework can serve multiple restaurants, each restaurant individually) or a simulator from which it receives real-time data (or historical data in the case of the simulator) about vehicles and customers. It provides decisions, on a real-time basis, that consist of information about which vehicle delivers food to what customer and in which order. TSB then forwards the information about new customer orders and about available and dispatched vehicles to the Decision-Making (DM) component that computes the routing plans. 

\subsection{Decision-Making Component}

As mentioned above, the DM component deals with the problem of assigning orders to the vehicles as well as planning the delivery paths for them. Algorithm~\ref{alg} summarizes the process. It initially monitors for customer orders that are almost ready (in the next five minutes), for orders that have been dispatched for delivery (i.e., vehicles with these orders have left the restaurant), and for vehicles that are about to return to the restaurant (in the next five minutes). Then, the set of customers $C$ and the set of vehicles $V$ are updated. According to the updated data about the customers and their locations, we update the graph $\mathcal{G}$, where $\mathit{time}$ and $\mathit{dist}$ functions (on the graph's edges) can be determined, for instance, by leveraging OSMNX, a library of OpenStreetMaps\texttrademark~ (OSM). Note that we do not explicitly consider vehicles' capacity since the amount of orders that are delivered never exceeds or even gets close to the capacity of a given vehicle. 
Then, we construct a VRPTW task $T=(V,C,\mathcal{G})$ while setting the time windows $[0,\mathit{deadline(c)}]$ for each $c\in C$, where $\mathit{deadline}(c)$ represents the deadline for the customer $c$. Deadlines are set by the restaurant (our customer) according to its processes. We then proceed to solve $T$, initially trying to meet all the deadlines and, if the task is not solved within a given time limit, we keep increasing the deadline for all customers by $\delta$ until the VRPTW task is solved. Such an approach was designed to guarantee that the delay is kept within reasonable bounds for each customer (in contrast to optimising average delay, which might impose an intolerably large delay for some customers).

The rationale behind considering orders that are not yet ready is twofold. Firstly, it is useful to adjust for aspects such as the time needed to physically group the orders and load them into a delivery vehicle or the later arrival of drivers. Secondly, locations of delivery for some later orders might be close to the delivery locations of earlier orders which might allow more efficient routing. The five-minute threshold is given by the restaurant (our customer) as it aligns with its internal processes.
Decisions about when the vehicle is to be dispatched with the orders are made by the respective restaurant. Of course, by then all the orders have to be ready and the vehicle has to be at the restaurant.

\subsection{Technology-Specific Bridge}

\begin{figure*}[!t]
    \centering
    \includegraphics[width=0.8\textwidth]{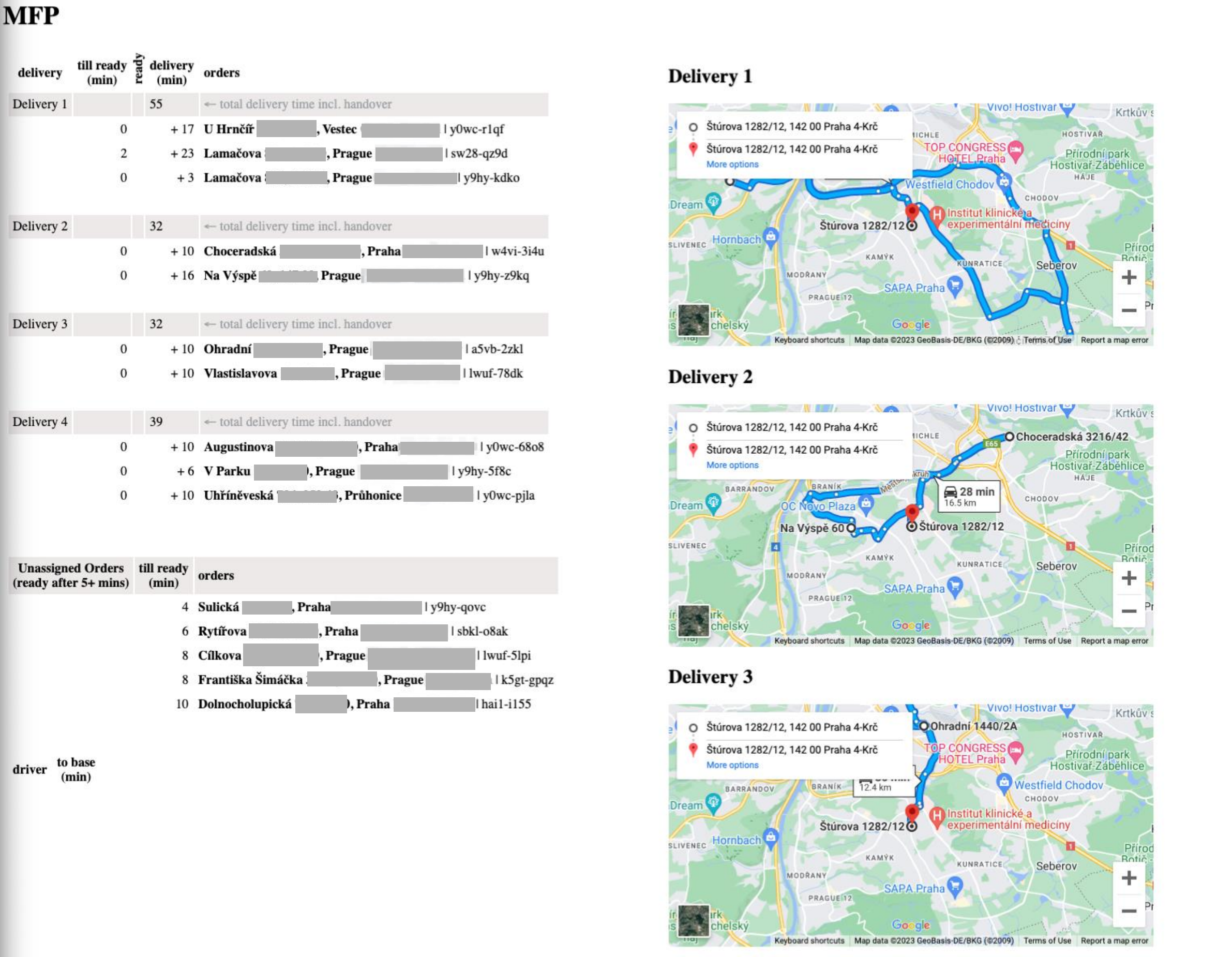}
    \caption{A screenshot of the recommender. On the left-hand side, the recommender shows the ``grouped'' deliveries (with not-yet-ready orders at the bottom). On the right-hand side, the recommender depicts the delivery routes in maps.}
    \label{fig:screenshot}
\end{figure*}

TSB, as mentioned before, can be understood as an interface between the restaurants (or the simulator) and the DM component. To interpret (and verify) the solutions of the VRPTW tasks (returned by the DM component), we translate these solutions into sequential plans that contain actions, specified in PDDL, that are then forwarded to restaurants (or the simulator). 

In particular, the PDDL model consists of six actions. Assigning customers' orders to the vehicles is done by actions \textsf{assign-order}, which assigns an order to a specific delivery, and \textsf{assign-delivery}, which assigns a delivery to a specific vehicle. Note that these two actions correspond to a process in restaurants, where the finished orders are initially grouped before being loaded into a vehicle for delivery. The delivery process starts with the \textsf{dispatch-delivery} action, \textsf{drive} and \textsf{deliver-order} are for driving to the delivery location and handing the delivery to the customer, and the \textsf{finish-delivery} action represents that the driver returned to the restaurant. Again, these actions correspond to the processes that the restaurants implement for the delivery process. 

These plans are then validated in UPF~\cite{upf}. Then, the current plans are displayed in a \emph{recommender}, which displays the groups of orders and visualises the delivery routes on a map (Google Maps\texttrademark), so that the restaurant staff can make a decision about when the orders should be loaded to a vehicle and dispatched for delivery (a screenshot of the recommender is shown in Figure~\ref{fig:screenshot}). 





\subsection{Simulator}

\begin{table}[!t]
    \centering
    \begin{tabular}{|c|c|c|c|c|c|}
    \hline
      DT   & DD & TD & PD & P10D  \\
      \hline
      1.08   & 1.09 & 0.67 & 0.90 & 0.61\\
      \hline
    \end{tabular}
    \caption{Average improvement of Cloud Kitchen against historical data (measured by ratio to historical data). In particular, we consider total driven time (DT), total driven distance (DD), total delay (TD), and the number of orders delivered after the deadline (PD) or 10 minutes after the deadline (P10D).}
    \label{tab:cummulative_results}
\end{table}

\begin{figure*}[!t]
    \centering
    \includegraphics[width=0.92\textwidth]{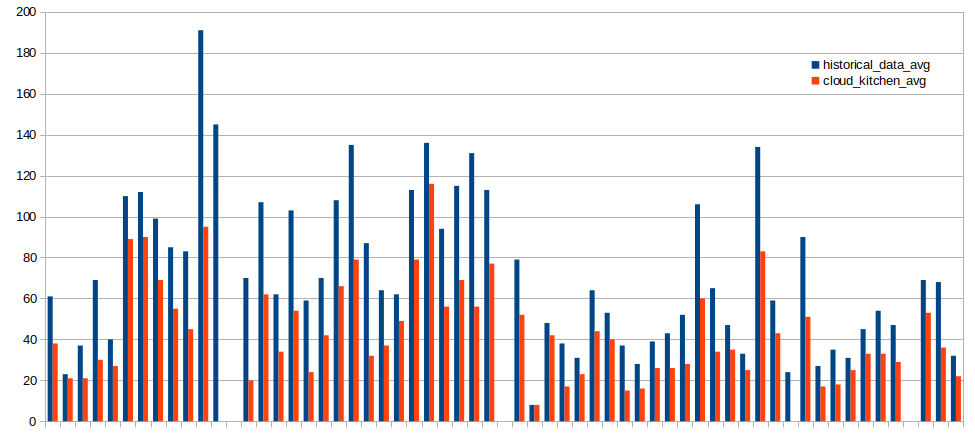}
    \caption{Comparison of the numbers of orders per day delivered later than 10 minutes ($y$-axis) in historical data and planned by the Cloud Kitchen platform. Particular days are on the $x$-axis.}
    \label{fig:results}
\end{figure*}

The purpose of the simulator is to provide a realistic estimate of the impact of the decisions of the Cloud Kitchen platform made on historical data and visualize them to the potential customers (restaurants with food delivery). The simulator tracks the status of the orders, i.e., when the order is received by the restaurant, when it is cooked, then assigned for delivery, dispatched, en route,  until it finally arrives to the customer who collects it. Similarly, we can track the state of each vehicle such as before delivering orders, the vehicle is ready (waiting at the restaurant), then the orders are assigned and loaded into it, and, after that, the vehicle goes to the customer to deliver the order, and after it delivers all the orders, it comes back to the restaurant.

Implementation-wise, the simulator is built on top of a process-based discrete event simulation framework named SimPy. The framework provides a linear, dimensionless timeline and allows for scheduling events to be executed at a specific time on this timeline. Events can then be processed, chained, or simply observed. One simulator tick represents one minute of real-time.

As mentioned above, driving distances and driving times are taken from OSM. To get a more realistic estimate of the actual driving time (in traffic conditions as they were in historical data), we have to determine a factor by which we multiply the estimated driving times from OSM. This is done by initially simulating the deliveries as they were arranged in the historical data and then by dividing driving times from the simulation by the driving time estimates from OSM.

Vehicle dispatching is automated such that if all orders in a batch (allocated to a single vehicle) are ready and there is some vehicle waiting at the restaurant, then the vehicle is dispatched with that batch of orders.  

The visual part contains an interactive map (based on Google Maps) consisting of pinpoints at the locations of customers that also indicate the time until the deadline and the current positions of the vehicles. On the right-hand side, there is a sidebar that shows the status of the orders. 

\section{Empirical Evaluation}


To evaluate our Cloud Kitchen platform, we have simulated decisions on the platform on a real historical dataset consisting of one restaurant, nine vehicles, and more than 14000 customers (orders) spanning 61 days. As a solver for the VRPTW tasks we used Google OR Tools\texttrademark, which we configured on the ``path\_cheapest\_arc'' option for the ``first\_solution\_strategy''. The reason for considering the ``first\_solution\_strategy'' is the need to operate near real-time. For that reason, we have also set the timeout for the solver to 50 milliseconds. Setting the ``path\_cheapest\_arc'' option was based on an empirical evaluation on a subset of our dataset. Also, we have found out that the VRPTW task has been solved within 30 milliseconds, or not at all (within 30 seconds), so, to give some leeway, we used the limit of 50 milliseconds per a VRPTW task. For the ``deadline extension'' loop (Lines~9--15 in Algorithm~\ref{alg}), we used a 1-second time limit (to keep near real-time reasoning). 

As mentioned above driving distances and driving times are determined from OSM. To adjust for the conditions that were at the time of the dataset, we initially simulated the deliveries as they were arranged in the historical data and then we compared driving times (from the simulation) against the driving time estimates from OSM. From that comparison, we determined a factor ($1.6666$) by which we multiplied all driving time estimates from OSM to provide realistic estimates for our experiments. We would like to note that in real-world scenarios we use HERE maps to provide current driving time estimates for the recommender as HERE maps are more precise than OSM, however, the number of routing calls in the simulator is about three orders of magnitude higher than in the (real-world) recommender, which would make HERE maps usage in simulations computationally expensive. 

The results, summarized in Table~\ref{tab:cummulative_results}, show that while driving time and driving distance slightly increased (by $8\%$ and $9\%$, respectively), the total delay has decreased by about one third, and the number of orders delivered later than 10 minutes after their deadline dropped by almost $40\%$. To give a better perspective, we provide, in Figure~\ref{fig:results}, day-by-day results comparing how many ``later than 10 minutes after deadline'' orders were in the historical dataset and were generated by the Cloud Kitchen platform, respectively. It can be seen that the numbers were not worse and often much better than in the historical dataset. There are two exceptions in days $12$ and $50$, where at least one planning episode failed (did not generate any routing plan within the time limit), and hence no results are shown in the simulator for that day as we could not validate the whole plan (for the day). In practice, if a planning episode fails, we can relax the deadlines to provide (at least) some recommendation for the restaurant.

From the results, we can see that the Cloud Kitchen platform has a strong potential to improve customer experience as it can considerably reduce the delay in order delivery. It has been confirmed by the restaurant (that provided us with the data) that orders delivered later than 10 minutes after their due time incur additional costs as the customers tend to ask for refunds and/or are less likely to order again. 
Despite the slight increase in costs (increased driving time and driving distance), minimising the delays beyond 10 minutes is hence much more beneficial for the restaurants.


\section{Conclusion}

In this paper, we have introduced the Cloud Kitchen platform that provides a decision-making tool for restaurants (or kitchens) with food delivery. In particular, the platform provides recommendations about which orders should be delivered together as well as providing routing for the vehicles (i.e., in which order the deliveries should be made). Such a problem can be modeled as VRPTW, which is tailored to address it. On the other hand, PDDL plans are better explainable to the restaurant staff who dispatch the orders as these plans are aligned with the processes in the restaurant. Hence, we translate solutions of VRPTW tasks into PDDL plans and leverage UPF to validate the plans before they are displayed to the restaurant staff in the recommender. The Cloud Kitchen platform also contains a simulator that can simulate the decisions (made by the platform) on (real) historical data. The purpose of the simulator is to show the advantages of using the platform for potential customers (restaurants).

We have empirically evaluated our platform on a real historical dataset, where we have minimized the number of delayed deliveries and the total delay. Although, as we showed, such an optimization might increase driving time and distance and thus incur additional costs, the number of delayed orders, especially those delayed by more than $10$ minutes, dropped considerably. Hence, the platform has the potential to improve customer satisfaction considerably, which has more value to the businesses despite the (slightly) increased costs of delivery (as confirmed by the restaurant).

In the future, we plan to extend our platform by considering different types of vehicles (e.g., scooters), distinguishing between different types of food (e.g., warm or cold food), and optimizing the process of preparing food in the kitchens.

\section*{Acknowledgements}

The work is co-funded by the AIPlan4EU project, which is funded by the European Commission – H2020 research and innovation programme under grant agreement No 101016442

\bibliography{aaai24}

\end{document}